\documentclass[graybox]{svmult}

\usepackage{mathptmx}       
\usepackage{helvet}         
\usepackage{courier}        
\usepackage{type1cm}        
\usepackage{makeidx}         
\usepackage{graphicx}        
\usepackage{multicol}        
\usepackage[bottom]{footmisc}
\usepackage{array, booktabs} 
\usepackage{multirow}
\usepackage{enumitem}
\usepackage{mdframed}
\usepackage{xcolor}
\usepackage{soul}

\newcolumntype{C}[1]{>{\centering\arraybackslash}p{#1}}
\newcolumntype{R}[1]{>{\raggedleft\arraybackslash}p{#1}}
\newcolumntype{L}[1]{>{\raggedright\arraybackslash}p{#1}}

\graphicspath{{./figs/}}

\makeindex             

\begin{document}

\title*{AI and Ethics --- Operationalising Responsible AI}
\author{Liming Zhu, Xiwei Xu, Qinghua Lu, Guido Governatori, Jon Whittle}
\institute{Liming Zhu, Xiwei Xu, Qinghua Lu, Guido Governatori, Jon Whittle\at CSIRO Data61, 13 Garden Street, Eveleigh NSW, Australia \\\email\{firstname.secondname\}@data61.csiro.au}
%
%
\maketitle

\abstract*{\index{trustworthiness} vs. \index{trusted}}

\abstract{In the last few years, AI continues demonstrating its positive impact on society while sometimes with ethically questionable consequences. Building and maintaining public trust in AI has been identified as the key to successful and sustainable innovation. This chapter discusses the challenges related to operationalising ethical AI principles and presents an integrated view that covers high-level ethical AI principles, general notion of trust/trustworthiness and product/process support in the context of responsible AI, which helps improve both trust and trustworthiness of AI for a wider set of stakeholders.}

\section{Introduction}
\label{sec:chapt_echics_introduction}

When it comes to AI and Ethics/Law\footnote{Law is usually considered to set the minimum standards of behaviour while ethics sets the maximum standards so we will use the word "ethics" throughout the chapter.}, there are two interrelated aspects of the topic. One is on how to design, develop, and validate AI technologies and systems responsibly (i.e., Responsible AI) so that we can adequately assure ethical and legal concerns, especially pertaining to human values. The other is the use of AI itself as a means to achieve the Responsible AI ends. In this chapter, we focus on the former issue.

In the last few years, AI continues demonstrating its positive impact on society while sometimes with ethically questionable consequences. Not doing AI responsibly is starting to have devastating effect on humanity, not only on data protection, privacy and bias but also on labour rights and climate justice~\cite{crawford2021hidden}. Building and maintaining public trust in AI has been identified as the key to successful and sustainable innovation~\cite{CDEI20}. Thus, the issue of ethical AI or responsible AI has gathered high-level attention. Nearly one hundred 
principles and guidelines for ethical AI have been issued by private companies, research institutions, and public organisations~\cite{jobin2019global,Dignum-book-reponsible-AI} and some consensus around high-level principles has emerged~\cite{fjeld2020principled}. On the other hand, principles and guidelines are far from ensuring the trustworthiness of AI systems~\cite{Mittelstadt19}. Complicating the issue further, humans and societies perceive trust in AI in intricate ways, which does not necessarily closely match the trustworthiness of a particular AI system~\cite{hidalgo2020humans, KPMG20-TrustinAI, Miller20191}.

The remainder of the paper is organized as follows: Section~\ref{sec:chapt_echics_challenges} discusses the challenges of existing works on ethical AI. The framework with an integrated view of three aspects of ethical AI is discussed in Section~\ref{sec:chapt_echics_solution}. Section~\ref{sec:chapt_echics_example} shares our experience and observations in a crop yield prediction project. Section~\ref{sec:chapt_echics_discussion} talks about the high-level ethical principles and their operationalization. Finally, section~\ref{sec:chapt_echics_summary} concludes the chapter.

\section{Ethical AI Challenges}
\label{sec:chapt_echics_challenges}

\subsection{Classification of Existing Works on Ethical AI}
Significant research has gone into addressing ethical AI challenges. In this section, we discuss the existing works, which fall into three large categories:

\begin{enumerate}
    \item \textbf{High-level ethical principle frameworks.} A large number of high-level ethical principle frameworks~\cite{jobin2019global} (e.g., Australian AI Ethics Principles\footnote{\url{https://www.industry.gov.au/data-and-publications/building-australias-artificial-intelligence-capability/ai-ethics-framework/ai-ethics-principles}}). They identify the important ethical and legal principles responsible AI technologies and systems are supposed to adhere to. Some effort, such as \cite{Shneiderman20}, further divides these high-level principles into guidelines at the team, organisational and industry level. These high-level principles are hard to operationalise for many reasons~\cite{Mittelstadt19} we will discuss later.
    \item \textbf{Ethical algorithms.} Significant research has gone into ethical algorithms where the formulation of some ethical/legal properties is amenable to mathematical definitions, analysis and theoretical guarantees. These include properties such as privacy~\cite{ji2014differential} and fairness~\cite{mehrabi2019survey} or for  specific types of AI systems~\cite{dong2020survey}. This covers mechanisms that deal with pre-processing of data (to remove bias or individualistic characteristics), the learning process itself (to take into consideration of ethical constraints), learned models (to be further compliant with ethical constraints) and predictive results (to correct for residual bias or revealing individualistic information). However, these mechanisms are algorithm focused with limited theoretical heuristic,   confined to a small number of quantification-amenable properties, and a small subset of ethical principles and human values~\cite{Whittle20-ICSE}. Most of the time, these ethical-aware algorithms are too complicated to explain to less numeracy-equipped stakeholders and not connected to the broader decision making process~\cite{CDEI20}. They are also not linked to the software development processes, especially system design methods, requirements engineering or user-centred design (UCD) processes.
    \item \textbf{Human values in software engineering and their operationalisation.} Recently, there has been emerging research in human values in software engineering and their operationalisation~\cite{hussain2020casestudy,hutchinson2020accountability}, including:
    \begin{enumerate}
        \item Extension of value-based design methods (e.g., value-sensitive design - VSD) \cite{VSD2015}
        \item Extension of human factor research on productivity and usability into human values consideration but still limited to a small subset of human values \cite{Whittle20-ICSE}
        \item Software engineering methods for embedding human values and ethical consideration throughout the software development life cycle (SDLC)~\cite{hussain2020casestudy, value-based-RE, value-first-se}
        \item Architecture and design patterns that can improve (qualitatively or quantitatively)~\cite{privacyDesign} or assure (with strong mathematical guarantees), ``by-design'', certain ethical or human-value related quality attributes such as privacy and non-maleficence (e.g. security, safety and integrity)~\cite{federated-learning-and-blockchain, lo2020systematic}
    \end{enumerate} 
\end{enumerate}

\subsection{Issues of Existing Works on Ethical AI}

We identify three issues in current research work regarding operationalising ethical principles to achieve the ultimate trust from stakeholders.

\subsubsection{Mixing Inherent Trustworthiness with Perceived Trust}

The inherent and technical \textit{``trustworthiness''} of an AI system can be directly reflected in technologies/\textbf{products} via code, algorithms, data or system design or indirectly reflected via the software development \textbf{processes}). On the other hand, trust is a stakeholder's (i.e., truster's) subjective estimation of the trustworthiness of the AI system. This subjective estimation is based on a truster's expected and preferred future behaviour of the AI system. Mixing the two in terms of identifying assurance mechanisms and presenting trustworthy evidence can overlook the additional and special mechanisms required to gain trust (different from the ones for gaining trustworthiness).

A highly technically trustworthy system may not be trusted by trusters for one reason or another, rationally or irrationally. This is because a truster's subjective estimation of the system's trustworthiness and expectations may have a significant gap compared to the system's inherent trustworthiness. It can also be the other way around when a truster overestimates a system's trustworthiness and puts undue trust into it.

The reasons for this gap may be related to several issues: 
\begin{itemize}
    \item a truster's numeracy (impacting the understanding of different types of trustworthiness evidence);
    \item a truster's prior beliefs and experiences;
    \item a truster's preferences and expectations on acceptable behaviours, types of evidence and explanation\cite{Miller20191};
    \item a system's observable behaviours to a truster.
\end{itemize}

\subsubsection{Operationalising Ethical Principles}

There are many reasons why we still lack systematic methods to operationalise the high-level ethical principles. Some are due to the AI field's relatively short history (e.g., compared with the medical field) thus lacking professional norms, legal and professional accountability mechanisms, clear common aims, fiduciary duties and importantly proven methods to translate principles into practices~\cite{Mittelstadt19}. Others are due to the lack of consideration of a wider set of human values~\cite{mougouei2018operationalize} such as political self-determination and data agency beyond technical dependencies~\cite{IEEE19-EthicallyAlignedDesign}. 

We believe another important factor is due to the relatively narrow attempt to operationalise human values and ethical principles into verifiable ``product'' trustworthiness (via mathematical guarantees) without systematically exploring a wider variety of mechanisms in development processes to improve both \textbf{trustworthiness} and \textbf{trust}. Looking at process mechanisms can include highly tailored evidence gathering and communication mechanisms for different types of trusters. These will help close the gap between their subjective estimation and the system's more objective inherent trustworthiness.

\subsubsection{Unique Characteristics of AI}

Finally, many of the works do not actively consider the unique characteristics of AI during operationalisation. Referring to one~\cite{AU-AI-Ethics-Principles} of the many definitions of AI, AI is a collection of interrelated technologies used to \textbf{solve problems autonomously}, and perform tasks to achieve defined objectives, in some cases \textbf{without explicit guidance from a human being}. AI has its own agency \cite{Dignum-book-reponsible-AI} reflected in its autonomy (i.e., acting independently), adaptability (i.e., learning in order to react flexibly to unforeseen changes in the environment) and interactivity (i.e., perceiving and interacting with other agencies, human or artificial). So by AI's definition and its inherent autonomy-related characteristics, it would be impossible (not simply hard) to accurately and completely specify all the goals, undesirable side-effects and constraints (including ethical ones) at its finest level of details. This is known as the value alignment problem: given an optimisation algorithm, how to make sure the optimisation of its objective function results in outcomes that we actually want, \textit{in all respects}? As one saying goes, ``\textit{It never does just what I want, but only what I tell it.}'' This inherent under-specification issue is both a boon and a bane of AI. Thus it is important not just to use guarantee mechanisms but to introduce a range of product and process-related risk mitigation mechanisms. This will include things like continuous validation and monitoring of systems~\cite{Staples16} after deployment, broadening specifications and real-world validation~\cite{damour2020underspecification}.

\section{Our Solution}
\label{sec:chapt_echics_solution}

Although previous work has produced high-level ethical AI principles, general notions of trust vs trustworthiness and product vs process support, they have not been integrated into the context of responsible AI. The contribution of this book chapter is the integrated view of the three aspects and how they help improve both trust and trustworthiness of AI for a wider set of stakeholders. This integrated view includes three components:

\begin{itemize}
     \item the difference between trust and trustworthiness in the context of ethical AI principles;
     \item how different product and process mechanisms can achieve trustworthiness for different ethical principles;
     \item how different product and process evidence can be presented to different types of trusters to improve the accuracy of their subjective estimation so they match the inherent trustworthiness of the systems.
\end{itemize}

\begin{figure}[t]
	\centering
	\includegraphics[width=0.95\columnwidth]{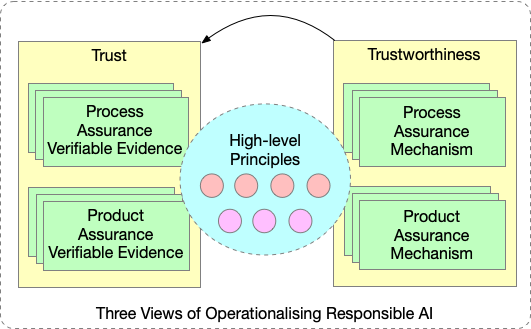}
	\caption{Conceptual Model.}
	\label{fig:conceptual}
\end{figure}

Our work also takes into consideration of the autonomy characteristics of AI and its inherent under-specification challenges. Figure~\ref{fig:conceptual} gives a graphical representation of our framework. 

For conceptualising the relationship between trust and trustworthiness, we use the definitions and concepts from Bauer's work~\cite{bauer2019conceptualizing}: 

\vspace{1em}
\begin{mdframed}[backgroundcolor=gray!15] 
    \textit{Trust $P_{{A_i}{t_o}}$ is truster $A_i$'s subjective estimate of the probability $P_{b_j}$ that $B_j$ will display $A_i$' preferred behavior $X_{k{t_1}}$ i.e., of $B_j$'s trustworthiness.}
\end{mdframed}
\vspace{1em}

In our work, $B_j$ represents an AI system. The behaviours $X_{k{t_1}}$ displayed by $B_j$ can include functional behaviours, behaviours that handle ethical constraints (e.g. privacy, security, reliability, safety and other human values and wellbeing) and meta-level behaviours (e.g.transparency, explainability~\cite{holzinger2019causability} and accountability). A truster's subjective estimate at $t_0$ is about the AI system's future behaviour at $t_1$. There are some arguments around the concept of trust being binary (rather than probabilistic) in reality as you either trust something or not. When you eventually decide to accept a system to be trusted, you then accept the associated harm if the trusted party fails. We believe this binary notion is consistent with the probabilistic notion if you introduce a thresh-hold along the probability spectrum to discern trusted or not trusted. 

We use Australia's ethical AI principles~\cite{AU-AI-Ethics-Principles} and their definitions as a close-enough representation of the many similar ones~\cite{jobin2019global,fjeld2020principled} around the world. Australia's ethical AI principles contain eight key principles.

\begin{itemize}
    \item \textit{P1: Human, social and environmental wellbeing:} Throughout their lifecycle, AI systems should benefit individuals, society and the environment.
    \item \textit{P2: Human-centred values:} Throughout their lifecycle, AI systems should respect human rights, diversity, and the autonomy of individuals.
    \item \textit{P3: Fairness:} Throughout their lifecycle, AI systems should be inclusive and accessible, and should not involve or result in unfair discrimination against individuals, communities or groups.
    \item \textit{P4: Privacy protection and security:} Throughout their lifecycle, AI systems should respect and uphold privacy rights and data protection, and ensure the security of data.
    \item \textit{P5: Reliability and safety:} Throughout their lifecycle, AI systems should reliably operate in accordance with their intended purpose.
    \item \textit{P6: Transparency and explainability:} There should be transparency and responsible disclosure to ensure people know when they are being significantly impacted by an AI system, and can find out when an AI system is engaging with them.
    \item \textit{P7: Contestability:} When an AI system significantly impacts a person, community, group or environment, there should be a timely process to allow people to challenge the use or output of the AI system.
    \item \textit{P8: Accountability:} Those responsible for the different phases of the AI system lifecycle should be identifiable and accountable for the outcomes of the AI systems, and human oversight of AI systems should be enabled.
\end{itemize}

\subsection{The difference Between Trust and Trustworthiness in the Context of Ethical Principles}

We group the eight principles in two categories based on their nature and characteristics. The first group includes the first five principles (P1--P5), which are human values and ethical constraints similar to the non-functional software qualities~\cite{iso2011iec25010} to be considered. The second group includes the last three principles (P6, P7 and P8), which are meta-level governance issues.

\subsubsection{Principles as Software Qualities}

These principles sometimes can be framed as functional requirements of a software system. If that is the case, in the context of software engineering, the methodology of requirement engineering could be adopted to ensure that the requirements captured are as accurate and complete as possible. It's worth noting again that AI systems can not be fully specified and will try to solve the problems \textbf{autonomously} with a level of independence and agency. On the other hand, there will be conflicting requirements whereby tradeoff decisions need to be made. 

Some principles, such as security, reliability and safety, are the non-functional properties well studied in the dependability research community~\cite{dependability}. These principles can be captured as non-functional requirements and considered from the early stage of system design. There are technical mechanisms or reusable design fragments, like patterns and tactics, that could be applied to fulfil the quality requirements~\cite{bass2003software}. Privacy is not a standard software quality~\cite{iso2011iec25010}, but has been treated as an increasingly important property of a software system to realise regulatory requirements, like GDPR (General Data Protection Regulation)\footnote{General Data Protection Regulation, https://gdpr-info.eu/.} and Australia Privacy Act, into technical artifacts. Reusable practices and patterns have been summarised in both industry and academia for privacy \cite{privacyDesign}. Fairness is a concern that the machine learning developers should consider from the early stage of the data processing pipeline. Similar to before, collections of best practice and mechanisms to remove bias at different stages of the pipeline have been compiled~\cite{mehrabi2019survey}. The reason to group these principles is that they can be handled and validated using a similar approach: the methodology of how non-functional properties are handled in software system design. Some principles can be validated in a quantifiable way, like reliability. Others could be validated against process-oriented best practices, methods and widely used patterns. 

\subsubsection{Principles as Governance Issues}

The principles within this group are largely governance issues and designed to improve truster's confidence in the AI system. They can be seen as clear requirements for certain functionality provided by the software system or entities providing the system. For example, contestability requires a system or entity function that allows the trusters to challenge the output or use of the AI system. Transparency and explainability are similar in that the users can have access to the system, data, algorithms to understand it, including receiving an explanation of a decision or prediction.

\subsubsection{Trust vs. Trustworthiness of Ethical AI Principles}

Each principle can have different implications in terms of trust vs. trustworthiness. Essentially, the difference is between what an AI system can objectively perform (trustworthiness) and what a truster/stakeholder “prefers/wants” (trust expectation) and their subjective estimation of the behaviour of the AI system. And it has been observed that human may have very different expectations of AI or human even their trustworthiness are similar ~\cite{hidalgo2020humans}. We list these differences in Table~\ref{tab:trust-vs-trustworthiness}.

\begin{table}[htbp]
   \centering
   \caption{Trust vs. Trustworthiness}\label{tab:trust-vs-trustworthiness}

   \begin{tabular}{C{0.2\linewidth}L{0.4\linewidth}L{0.4\linewidth}}
    \toprule
    & \textbf{Trustworthiness}      & \textbf{Trust}         \\
    \midrule
    Human, social and environmental wellbeing & whether stakeholder requirements are captured accurately and completely and tradeoff are made in an informed and consultative way & whether a truster perceives the entity, the process and society safeguards \cite{KPMG20-TrustinAI} collecting requirements and making tradeoffs are trustworthy and whether a truster's requirements are addressed adequately.\vspace{0.5em}\\
    \cmidrule(l){1-3}
    Human-centred values & whether comprehensive sets of relevant values are considered \cite{Whittle19} throughout the SDLC & whether a truster perceives the entity, the process and society safeguards \cite{KPMG20-TrustinAI} of developing, verifying and validating the system are trustworthy.\vspace{0.5em}\\
    \cmidrule(l){1-3}
    Fairness & \multirow{3}{*}{\parbox{4.2cm}{whether data, learning algorithms, learned models, decisions, predictive results, and overall designs are developed with quantifiable fairness, privacy and security constraints in mind and satisfy the reliability and safety requirements.}} &  \multirow{3}{*}{\parbox{4.2cm}{whether a truster understands how these constraints are satisfied and perceives the entity, the process and society safeguards of developing, verifying and validating the system are trustworthy.}}\\
    \cmidrule(l){1-1}
    Privacy protection and security & & \\
    \cmidrule(l){1-1}
    Reliability and safety \vspace{2em} & & \\
    \cmidrule(l){1-3}
    Transparency and explainability & whether system requirements, specifications, data, algorithms, models, decisions, system designs and source code, i.e., all related artifacts, are open for stakeholder inspection and understandable. & whether a truster perceives they can understand the artifacts and explanation associated or they can delegate such access and understanding to a trustworthy third party or society's safeguard regulations.\vspace{0.5em}\\
    \cmidrule(l){1-3}
    Contestability & whether there is a timely process specified for people to challenge the use or output of the AI system at an individual decision, group or society level. & whether a truster perceives there is a timely and trustworthy process run by trustworthy entities to challenge the use or output of the AI system at an individual decision, group or society level.\vspace{0.5em}\\
    \cmidrule(l){1-3}
    Accountability & whether there are entities and humans identified to be accountable for the outcomes of the AI systems. & whether a truster perceives the identified accountable entities and humans are the right ones and can bear proportionate responsibilities under adverse outcomes.\\
    \bottomrule
   \end{tabular}
\end{table}

As we can see, trust is about subjective estimation and perception and often not limited to the AI system's trustworthiness properties (whether via software artifacts or development processes). As identified in \cite{KPMG20-TrustinAI},
multiple factors contributes to people's trust in an AI system, including factors not related to a specific AI system such as current society safeguards such as regulations, overall AI uncertainty, job impact and familiarity of AI. Some factors may be related to the organisations that build the AI systems, use the AI system or evaluate the AI system. The specific characteristic of the AI system only plays a minor role in trust.

\subsection{Product and Process Mechanisms for Trustworthiness}

We continue using the same grouping as the last section to analyse the eight principles to demonstrate different product and development process (including the people aspect) mechanisms to improve trustworthiness. This differentiation teases out a broader set of considerations to improve trustworthiness, which subsequently helps understand what are the different ways in which these mechanisms could be communicated to different stakeholders/trusters to improve trust. 

For human, social and environment wellbeing and human-centred values, we also consider organisational culture and SDLC methods including roles and agile practices. For quality attributes related ethical principles, we apply architecture and design processes, patterns and tactics and refer to a range of data, model \cite{damour2020underspecification} and algorithmic considerations. \cite{Roth-The-Ethical-Algorithm}. We consider design process (``in design''), design artifacts (``by-design'') and  designers (``for designers''). \cite{Dignum-book-reponsible-AI} For governance issues, we build upon the principles in \cite{Dignum-book-reponsible-AI, IEEE19-EthicallyAlignedDesign, Blackman20,hutchinson2020accountability}. 

We introduce some considerations and examples across the Product and Process/People dimension in Table \ref{tab:product-and-process}.
\begin{table}[htbp]
   \centering
   \caption{Product and Process Assurance Mechanisms}\label{tab:product-and-process}

   \begin{tabular}{C{0.2\linewidth}L{0.4\linewidth}L{0.4\linewidth}}
    \toprule
    & \textbf{Product}      & \textbf{Process and People}         \\
    \midrule
    Human, social and environmental wellbeing & Wellbeing metrics and associated requirements and subsequent design and continuous validation and monitoring features. Misuse cases and entities, undesirable side effects & Stakeholder engagement, independent boards and conflict/tradeoff resolution process. Roles, ceremonies and organisational culture. \vspace{0.5em}\\
    \cmidrule(l){1-3}
    Human-centred values & digital sovereignty \cite{IEEE19-EthicallyAlignedDesign}, culture norms, value statement and stories, value definition and explicit tradeoffs, value-aware training data, learning algorithms, models and decisions/predictions, continuous validation and monitoring & value tradeoff processes (with accuracy, profit and among different conflicting values). Roles, ceremonies and org culture. \vspace{0.5em}\\
    \cmidrule(l){1-3}
    Fairness & fairness definition and explicit tradeoffs, fairness-aware data, learning algorithms, models and decisions/predictions, continuous validation and monitoring, game-theoretical approach & fairness tradeoff processes (with accuracy, profit and among different conflicting fairness measures) roles, ceremonies, org culture, org-level boards, licensed developers. \vspace{0.5em}\\
    \cmidrule(l){1-3}
    Privacy protection and security & privacy and security requirements, privacy/security-by-design architectures (such as federated learning), privacy-enhancing data treatment and learning algorithms (such as differential privacy, secure-multiparty-computation), continuous validation and monitoring & privacy and security definition and tradeoff processes (with accuracy, profit and among different conflicting values). Roles, ceremonies, org culture. \vspace{0.5em}\\
    \cmidrule(l){1-3}
    Reliability and safety & reliability and safety requirements, reliability/safety architectures and designs patterns, reliability/safety-enhancing algorithms & reliability and safety tradeoff processes (with other quality attributes). Roles, ceremonies, org culture. \vspace{0.5em}\\
    \cmidrule(l){1-3}
    Transparency and explainability & Features that generate human understandable explanations tailored for different stakeholders. Registration and record keeping, provenance and documentation of all artifacts. Explanations of data, algorithm, models and decisions. Adjust model complexity. Have ``Why did you do that button''. & open process across SDLC including full access to artifacts from stakeholders or 3rd parties representing stakeholders. Processes for iterative exploration and explanation. Consider ``no algorithm allowed''. \vspace{0.5em}\\
    \cmidrule(l){1-3}
    Contestability & Features that allow human intervention of decisions ex ante and ex post.& Contestability definition. Roles, ceremonies, org culture,process conformance to standard processes\vspace{0.5em}\\
    \cmidrule(l){1-3}
    Accountability & Features that allow provenance and traceability of artifacts and decisions that always have a clear accountable entity and human. & Artifact specific life cycle management process (such as data accountability\cite{hutchinson2020accountability} and process creation).\\
    \bottomrule
   \end{tabular}
\end{table}
Many of the mechanisms presented in the table rely on industry-wide and society-wide work, in particular:
\begin{itemize}
    \item Multi-stakeholder sector-specific~\cite{CDEI20} and domain-specific guidelines (such as banned practices in digital platforms~\cite{EC19}, regulating software-based medical devices \cite{tga}, technical solutions and empirical knowledge bases~\cite{Mittelstadt19}).
    \item AI and data ethics board or other governance mechanisms at the organisation level~\cite{Blackman20} to oversee the overall AI-driven decision-making processes (not just algorithms and products).
    \item Regulator levers that incentivise organisations and create a level playing field for ethical innovation~\cite{CDEI20} including validation and certification agencies.
    \item Technical and non-technical ethics and human rights training for different roles and organisational awareness.
    \item Incentives for employees to play a role in identifying AI ethical risks~\cite{Blackman20}.
    \item Ongoing monitoring and engagement with stakeholders \cite{Staples16, Blackman20}.
\end{itemize}

Due to the under-specification and value alignment problem of AI, none of the assurance mechanisms can guarantee a desirable outcome alone. Each helps reduce the risk and the ongoing monitoring and engagement post-deployment plays a critical role in identifying and mitigating undesirable side effects early. 

\subsection{How to Present Trustworthiness Evidence to Different Types of Trusters}

It is important that we present trustworthiness evidence (e.g. product artifacts assurances and process/people assurances) to different types of trusters to help with their subjective estimation so the estimation matches the inherent trustworthiness of the AI systems. Here we have to take into considerations of two different aspects. 

\medskip
\noindent\textbf{Truster Preference}:
Based on expectation and past experiences, different types of trusters, such as AI algorithm developers, AI system developers, professional users of AI systems, affected subjects (whereby AI systems have impacts on them),  regulators and certifiers, or the general public may have different preferences of an AI system's behaviour.  

\medskip
\noindent\textbf{Truster-Specific Verifiable Evidence}: Different types of trusters may expect different types of evidence for assuring the trustworthiness of an AI system. They may have different abilities to understand and assess the evidence and expect different types of explanations \cite{Miller20191}. For example, an AI expert can assess the algorithms while a general public would have no use of the algorithms. A system developer may understand an AI algorithm and the data associated but have limited ability to evaluate the algorithm and bias in data.

\medskip 

When presenting evidence, such as assurance mechanisms in the last section, the following factors should be taken into consideration:

\begin{itemize}
    \item Consider a stakeholder's technical abilities to understand the evidence.
    \item Consider a stakeholder's resources and time to assess the evidence.
    \item Allow both stakeholders and 3rd parties who represent the stakeholders to examine the evidence.
    \item Allow broader ethical and legal "standing" so the evidence can be produced and examined at an individual decision, group and society level by a wide range of stakeholders or their delegates.
    \item Present both process and product assurance mechanisms to the right stakeholders in the right ways.
    \item Focus more on process mechanisms when ethical principles can not be easily defined and quantified.
    \item Focus more on the product mechanisms when ethical principles can be defined and quantified and technically assured.
    \item Improve overall awareness/familiarity of AI, and understanding of broader AI issues such as current society safeguards, AI's overall uncertainty and job impact.
\end{itemize}

\section{Example: Crop Yield Prediction Project}
\label{sec:chapt_echics_example}

In this section, we share our experience and observations in a crop yield prediction project. In the project, we applied a data-driven approach using machine learning algorithms to develop an improved version of a crop yield prediction model. The previous model was built using a domain model-driven approach and integrated into a commercialised software product. The project combines commercial satellite data and data collected from different farms, largely owned by individual farmers. The resulting machine learning model predicts future yields of farms in both the farm data collection region but also in new regions significantly different from the original regions. There are three types of stakeholders in the project: the technical team (a research organisation playing the role of both AI system developer and operation-time learning coordinator), data contributor (i.e., participating farmers who provided the data), and model user (i.e., farmers who use the models for yield prediction). We used a federated machine learning approach to deal with the data ownership problems and the non-IID (independent and identically distributed) data distribution problem. The learning coordinator (i.e., the technical team) designs and operates the model training process on multiple, distributed data contributors. The model was first trained locally on a local server and then sent to a central server for aggregating and improving a global model. The model user performs inference using the aggregated global model. Trustworthiness and trust issues manifested differently for different stakeholders. We applied our approach retrospectively to the project and made the following observations in term of trust and trustworthiness.

\begin{itemize}
  \item \textbf{Human, social and environmental wellbeing} 
    \begin{itemize}
    \item Positive wellbeing requirements were considered for the project itself but not for wider issues outside the project. 
    \item Potential misuse of data collected was considered but not for the predictive models.
    \item Both data contributors and model users have high-level trust in the technical team as an entity but have very little trust in any third parties who may also use the data (even for claimed wellbeing improvement reasons).
    \item Wellbeing requirements were specified as high-level goals, not quantifiable metrics.
    \end{itemize}

  \item \textbf{Human-centred values} 
    \begin{itemize}
    \item Human factors on usability and accessibility were considered, consulted and designed in the original prediction app. 
    \item No additional human-centred values were captured apart from the explicit ones below.
    \end{itemize}

\item \textbf{Fairness} 
    \begin{itemize}
    \item Fairness issues were considered at the training data, learning algorithm and model level but not explicitly explained to data contributors and model users. 
    \item A range of product assurance mechanisms was used, such as:
        \begin{itemize}
        \item Counterfactual analysis to discover potential hidden or proxy variables that lead to different yields.
        \item Sensitive-variable-aware data pre-processing including random split crop separation (wheat vs. barley), location separation (at paddock and region level). Each attribute may have associations, thus becoming a proxy variable regarding sensitive and protected attributes. 
        \end{itemize}
   \item Two team members built fairness-aware models in parallel as a process assurance.
   \item Used domain attributes that were understandable to farmers (data contributors and model users) in both training and explanation so farmers can have a higher level trust in the model regarding fairness across groups (not just accuracy).
    \end{itemize}

    \item \textbf{Privacy and Security} 
    \begin{itemize}
    \item The personal privacy of individual farmers and privacy/confidential info of farms were both identified as key requirements. 
    \item Potential misuses and harms of privacy info leakage were identified.
    \item Both data contributors and model users have high-level trust in the technical team as an entity.
    \item Federated learning, including strong privacy guarantees, was used as an architectural pattern fulfilling data privacy requirements. 
    \item The exchanged model updates was encrypted to improve model security. A model registry was established locally to maintain the mappings of encrypted models to the decrypted models.
    \item The concept of federated learning,  privacy guarantee and security mechanisms were difficult to explain to farmers, but the high-level trust in the project team coupled with the notion that their data never left the local server and strong encryption was used gained adequate trust. 
    \end{itemize}
    
    \item \textbf{Transparency and Explainability} 
    \begin{itemize}
    \item Transparency and explainability requirements were identified in the project.
    \item Although the farmers would not be able to evaluate data, code and models directly, the project team nevertheless made all artifacts open to stakeholders (under privacy and confidentiality constraints). 
    \item Code, model and data provenance were captured in Git and Bitbucket. 
    \item On the data contributor side, a data contributor registry was built to store and manage the information of the participating farmers and their paddocks that joined the learning process. Further, a model versioning registry was implemented to keep track of all local model versions of each data contributor and the corresponding global model. Both of the design mechanisms helped increase the trust of both model user and learning coordinator on the participating data contributors. 
    \item On the learning coordinator side, a decentralised aggregator was designed to replace the central server of the learning coordinator, which could be a possible single point of failure. This improved the trust of farmers (i.e., both data contributors and model users) on the learning coordinator.
    \end{itemize} 
    
     \item \textbf{Contestability} 
    \begin{itemize}
    \item Data contributors can always withdraw from the projects with their data deleted. However, the model trained by their data will not be immediately updated to remove the data. 
    \item Model users always have the right not to use the prediction or allow the prediction to be used by others.
    \end{itemize}
    
    \item \textbf{Accountability } 
    \begin{itemize}
    \item The accountability was largely governed by the legal agreement between data contributors, model users and the project team. 
    \item No role-level accountability was established, but the provenance of data, model and code allowed accountability to be examined.
    \end{itemize}


\end{itemize} 

\section{Discussion}
\label{sec:chapt_echics_discussion}
\subsection{Interpretation of high-level ethical principles}
Mappings of different ethical principles and guidelines have been recently studied in \cite{jobin2019global, fjeld2020principled}. Although a global consensus emerges on the core principles (e.g. privacy, transparency, fairness), there have been debates on the classification and definition of AI ethical principles. As we are using Australia's AI ethical principles, we discuss our interpretation and observations of these principles.
\begin{itemize}
\item \textbf{Autonomy.} Autonomy is not discussed explicitly in the framework. Instead, a broader principle about human-centred values is given, which covers human rights, diversity, and the autonomy of individuals. Autonomy refers to the freedom of AI system users to a range of activities, including self-sovereignty/determination, the establishment of relationships with others, selection of a preferred platform/technology, experimentation, surveillance, and manipulation~\cite{jobin2019global}. This might conflict with AI systems' inherent autonomy which refers to independent actions without a human being. Further, users (i.e., data owners) expect to exert full control over their data and activities without having to rely on others. To ensure autonomy, a distributed ledger technology like blockchain can play a vital role in the design and implementation of a decentralised AI platform, in which secure and self-determined interactions between stakeholders are enabled without a central coordinator. 

\item \textbf{Explainability.} Explainability is listed together with transparency in Australia's AI ethical principles. Transparency refers to disclosure of all related AI artifacts, such as source code, data, algorithms, models, and documentation. Although transparency enables explainability, explainability is more than helping stakeholders understand how an AI system works through responsible disclosure. Stakeholders expect to understand the reasons for AI system behaviour and insights about the causes of decisions. However, it is challenging to present the explanations (e.g., representation of data in a network - roles of layers/neurons) to gain the trust of stakeholders \cite{Explanations18, Miller20191} .

\item \textbf{Accountability.} Accountability principles are covered by most of the ethical AI principles and guidelines, including Australia's AI ethical principles. Ethical AI is often present together with responsible AI. However, the difference between responsibility and accountability is rarely discussed. Responsibility refers to the duty to complete a task throughout the lifecycle of an AI system. For example, a developer may be responsible for implementing an algorithm in an AI project. Accountability is the duty to be accountable for a task after it is completed, which happens after a situation occurs. For example, an AI startup company's CEO may be accountable for the inaccurate or biased decisions made by their AI system product and has to take the role to explain how the decision is made. Role-level accountable entities and humans should be clearly identified in AI projects. Product features and management processes that ensure provenance and traceability of AI artifacts and decisions can increase accountability of AI systems.    
\end{itemize}

\subsection{Operationalising AI ethics}
AI ethics can be operationalised in a variety of different ways. There are more approaches and mechanisms beyond high-level principles and low-level algorithms.

\begin{itemize}
\item \textbf{Requirements.} In addition to conventional functional requirements and non-functional requirements, ethical principles can be defined as a subset of requirements of AI systems. As discussed in Section 3.1, the eight principles can be classified into three groups:
1) P1 and P2 principles as functional requirements, 2) P3, P4 and P5 as non-functional requirements (software qualities), 3) P6, P7 and P8 as meta-level governance-related functional requirements to improve truster's confidence. However, it is hard to justify whether the P1 and P2 (i.e., human, social and environmental well being and human-centred values) are fulfilled adequately. Risk mitigation mechanism might be needed to deal with AI autonomy and ensure the fulfilment of P1 and P2.

\item \textbf{Design.} Design patterns/mechanisms can be proposed to address the ethical principles. Existing principles intend to protect users and the external world of AI system users. In distributed learning systems, trust issues also exist in between participating nodes. For example, in federated learning systems, learning coordinator might become a single point of failure. Thus, all the stakeholders should be carefully considered in design patterns/mechanisms.

\item \textbf{Operations.} Continuous validation and monitoring mechanisms are needed to check the ethical principles continuously. Engaged stakeholders need to identify the threshold for the significant impact which triggers the validation mechanisms. 

\item \textbf{Governance.}
\begin{itemize}

\item \textbf{Ethical maturity certification.} An AI ethics maturity certification scheme/ system could be developed to assess an organisation’s ethical maturity of AI project management. Based on the review of AI projects, different levels of ethical maturity certificates could be issued to an organisation. The level/type of certificate could be upgraded later.

\item \textbf{Ethics review.} Internal and external reviews on ethical impact can be conducted to address ethical principles. Representatives of stakeholders are expected to join the reviews.

\item \textbf{Project team.} When the development team is set up, team members’ diversity (e.g., background, cultures, and disciplines) should be considered. 
\end{itemize}
\end{itemize}

\section{Summary}
\label{sec:chapt_echics_summary}
In this chapter, we explored the interaction of three issues related to humanity and AI: hard-to-operationalise ethical AI principles, general notion of trust vs trustworthiness and product vs process support for trust/trustworthiness. We provided an integrated view of them in the context of AI ethical principles and responsible AI. It points to additional mechanisms especially process mechanisms and trust-enhancing mechanisms for different stakeholders. By using the example of crop yield prediction involving different types of data and stakeholders, we elicited the missing elements in operationalising ethical AI principles and potential solutions. We envision some future directions in ethical AI such as quantifying trust and its link with trustworthiness and novel process mechanisms for improving trust and trustworthiness. 

\bibliographystyle{spmpsci}
\bibliography{ref}

\end{document}